\documentclass{article}

\usepackage[preprint]{neurips_2022}

\usepackage[utf8]{inputenc} 
\usepackage[T1]{fontenc}    
\usepackage{hyperref}       
\usepackage{url}            
\usepackage{booktabs}       
\usepackage{amsfonts}       
\usepackage{nicefrac}      
\usepackage{microtype}      
\usepackage{xcolor}       
\usepackage{bm}
\usepackage{algorithm}
\usepackage{algpseudocode}

\usepackage{comment}

\usepackage{graphicx}
\usepackage{wrapfig}
\usepackage{mwe}
\usepackage{tikzducks}
\usepackage{babel}
\input{insbox}

\usepackage{mathtools,amsfonts,amssymb,amsthm}

\title{Importance Sampling for Stochastic Gradient Descent in Deep Neural Networks}

\author{%
  Thibault~Lahire \\
  Dassault Aviation\\
  Saint-Cloud, France \\
  \texttt{thibaultlahire.research@gmail.com} \\
}

\begin{document}

\maketitle

\begin{abstract}
Stochastic gradient descent samples uniformly the training set to build an unbiased gradient estimate with a limited number of samples. However, at a given step of the training process, some data are more helpful than others to continue learning. Importance sampling for training deep neural networks has been widely studied to propose sampling schemes yielding better performance than the uniform sampling scheme. After recalling the theory of importance sampling for deep learning, this paper reviews the challenges inherent to this research area. In particular, we propose a metric allowing the assessment of the quality of a given sampling scheme; and we study the interplay between the sampling scheme and the optimizer used. 
\end{abstract}

\section{Introduction}
\label{sec:intro}

Deep neural networks tend to become very large, with a high number of weights to optimize. Training deep neural networks is usually done with Stochastic Gradient Descent based algorithms \citep[SGD]{robbins1951stochastic}, such as RMSProp or ADAM \citep{kingma2015ADAM}. With such large architectures, training deep neural networks is computationally costly, since the cost of computing gradients is proportional to the number of weights. To alleviate this computational cost, one possibility is to take better SGD steps, so that the optimum is reached in fewer SGD iterations. 

SGD builds at each iteration an unbiased estimate of the empirical gradient by sampling uniformly the training set. However, at a given step of the learning process, some parts of the training set might be well handled by the neural network, whereas the error it makes on other parts is large. Some training data are more useful than others for training the neural network. Hence, sampling uniformly the training set at each step of the learning process to perform SGD might be sub-optimal. 

Importance sampling \citep{rubinstein2016simulation} applied to SGD has already been explored and applied to a wide range of datasets in supervised learning \citep{Needell2014SGD, wang2017accelerating}, as well as in reinforcement learning \citep{Schaul2015prioritized}. It gives sampling schemes for SGD which speed up convergence to the optimum in theory, by selecting the most helpful training data for the learning process. However, the optimal sampling scheme, yielding the highest convergence speed, is intractable in practice. To the best of our knowledge, all the works taking the best of importance sampling for SGD steps use approximations of the optimal sampling scheme.

\looseness=-1
This work aims at filling the gap between what is done in practice and the theory. The first contribution is a metric to assess the quality of a given sampling scheme. It allows to verify that a proposed sampling scheme is better (in terms of theoretical convergence speed) than the uniform sampling scheme. Then, we study the interplay between importance sampling schemes and optimizers. Indeed, one limitation of the works done in importance sampling for deep learning is the focus on the standard form of SGD. However, optimizers as RMSProp or ADAM are used in practice, and their update equation is different from the one of standard SGD. We show that the optimal sampling scheme for RMSProp and ADAM is different from the one for standard SGD. This has implications described in this paper.

This paper is structured as follows. The next section summarizes past contributions in importance sampling for SGD. Section~\ref{sec:contrib} introduces our metric assessing the quality of sampling schemes compared to the uniform sampling scheme. It also studies the interplay between importance sampling schemes and optimizers. Section~\ref{sec:exp} verifies experimentally the metric introduced and illustrates the interplay between sampling schemes and optimizers. Section~\ref{sec:conclusion} summarizes the contributions, highlights good practices to follow when using importance sampling for deep learning and concludes. 

\section{Background}
\label{sec:background}

\subsection{Importance sampling in deep learning}
\label{subsec:isdeep}

Let $\Psi(\theta, \cdot)$ be any deep neural network parameterized by $\theta$ and $\ell$ be the loss to minimize during training. Over a training set of $N$ items $(x_i, y_i)_{1 \le i \le N}$, the goal of training is to find: 
\begin{equation*}
    \theta^* \in \arg \min_\theta \frac{1}{N} \sum_{i=1}^N \ell (\Psi(\theta,x_i), y_i)
\end{equation*}
Writing $u$ the uniform probability distribution over the $N$ items of the training set, i.e. $\forall i \in [1;N], u_i = 1/N$, the empirical gradient of the loss function defined above can be written as an expectation:
\begin{equation*}
 \frac{1}{N} \sum_{i=1}^N \nabla_\theta \ell(\Psi(\theta,x_i), y_i) = \sum_{i=1}^N u_i \nabla_\theta \ell(\Psi(\theta,x_i), y_i) = \mathbb{E}_{i \sim u} [ \nabla_\theta \ell(\Psi(\theta,x_i), y_i)],
\end{equation*}
and the following unbiased gradient estimate can be built: 
$$\mathbb{E}_{i \sim u} [\nabla_\theta \ell(\Psi(\theta,x_i), y_i)] \approx \frac{1}{B} \sum_{i=1}^B \nabla_\theta \ell(\Psi(\theta,x_i), y_i) \ , \quad i\sim u$$
with $B$ the mini-batch size, $B \ll N$.
Writing $p$ any probability distribution over the training set, importance sampling can be used: 
$$\mathbb{E}_{i \sim u} [ \nabla_\theta \ell(\Psi(\theta,x_i), y_i)] = \mathbb{E}_{i \sim p} \left[ \nabla_\theta \ell(\Psi(\theta,x_i), y_i)\frac{u_i}{p_i}\right] = \frac{1}{N} \mathbb{E}_{i \sim p} \left[ \frac{1}{p_i} \nabla_\theta \ell(\Psi(\theta,x_i), y_i)\right].$$
The last expectation can be approximated with a mean estimator, just as we did to build the previous unbiased gradient estimate: 
$$\mathbb{E}_{i \sim p} \left[\frac{1}{p_i}\nabla_\theta \ell(\Psi(\theta,x_i), y_i)\right] \approx \frac{1}{B} \sum_{i=1}^B \frac{1}{p_i} \nabla_\theta \ell(\Psi(\theta, x_i), y_i)  \ , \quad i\sim p.$$ 
We introduce $G_i^{(t)} = w_i \nabla_{\theta} \ell(\Psi(\theta_t,x_i), y_i)$ with $w_i = 1/(Np_i)$ for any sampling scheme $p$ such that $\forall i \in [1;N], \quad p_i > 0$. 
This quantity will be useful to study SGD based algorithms using sampling $p$, possibly non uniform. Note that, when $p=u$, $w_i = 1$. 

\looseness=-1
Setting $\eta$ as a constant learning rate, a standard (full) gradient descent update has the form: $\theta_{t+1} = \theta_t - \eta \mathbb{E}_{i \sim p}[G_{i}^{(t)}]$. This can be surprising at first sight since it seems to depend on the sampling scheme $p$ used, whereas there is no sampling for the standard (full) gradient descent. It is indeed the case:  
\begin{equation}
    \mathbb{E}_{i \sim p}[G_{i}^{(t)}] = \sum_{i=1}^N p_i G_i^{(t)} =   \frac{1}{N} \sum_{i=1}^N \nabla_{\theta} \ell(\Psi(\theta_t,x_i), y_i),
\label{eq:const}
\end{equation}
and this notation gives consistency when writing the stochastic gradient descent update under sampling scheme $p$:  
\begin{equation}
\theta_{t+1} = \theta_t - \eta \frac{1}{B} \sum_{i=1}^B G_i^{(t)} = \theta_t - \eta \frac{1}{B} \sum_{i=1}^B \frac{1}{N p_i}\nabla_{\theta} \ell(\Psi(\theta_t,x_i), y_i).   
\label{eq:sgd_non_unif}
\end{equation}

Following the notations of \citet{wang2017accelerating}, let us define the convergence speed $S$ of SGD under a sampling scheme $p$ as $S(p) = -\mathbb{E}_{i \sim p} \left[ \|\theta_{t+1} - \theta^*\|_2^2 - \|\theta_t - \theta^*\|_2^2 \right].$
We recall that a stochastic gradient descent update with $B=1$ has the form $\theta_{t+1} = \theta_t - \eta G_i^{(t)}$, where $G_i^{(t)}$ is the gradient estimate built from sampling element $i$ with probability $p$.
The following derivations from \citep{wang2017accelerating} shed light on the relationship between variance of the stochastic gradient estimate and convergence speed:
\begin{align*}
S(p) &= -\mathbb{E}_{i \sim p} \left[ \theta_{t+1}^T \theta_{t+1} - 2 \theta_{t+1}^T \theta^* -  \theta_{t}^T \theta_{t} +2 \theta_{t}^T \theta^* \right] \\
&= -\mathbb{E}_{i \sim p} \left[ (\theta_t - \eta G_i^{(t)})^T (\theta_t - \eta G_i^{(t)}) + 2\eta {G_i^{(t)}}^T \theta^* - \theta_t^T \theta_t \right] \\
&= 2 \eta (\theta_t - \theta^*)^T \mathbb{E}_{i \sim p}[G_i^{(t)}] - \eta^2 \mathbb{E}_{i \sim p} [{G_i^{(t)}}^T G_i^{(t)}].
\end{align*}
Indeed, the term $\mathbb{E}_{i \sim p} [{G_i^{(t)}}^T G_i^{(t)}]$ can be called \emph{variance of the stochastic gradient estimate}, since it is linked to the covariance matrix $\mathbb{V}\textnormal{ar}_{i \sim p} [G_i^{(t)}]$ by $\mathbb{E}_{i \sim p} [{G_i^{(t)}}^T G_i^{(t)}] = \textnormal{Tr}( \mathbb{V}\textnormal{ar}_{i \sim p} [G_i^{(t)}]) + \mathbb{E}_{i \sim p} [G_i^{(t)}]^T \mathbb{E}_{i \sim p} [G_i^{(t)}]$. Recall also from Eq.~\ref{eq:const} that $\mathbb{E}_{i \sim p}[G_{i}^{(t)}]$ is a constant with respect to $p$. Hence, it is possible to gain a speed-up by sampling from the distribution that minimizes $\mathbb{E}_{i \sim p} [{G_i^{(t)}}^T G_i^{(t)}]$. 

\looseness=-1
Since minimizing $\mathbb{E}_{i \sim p} [{G_i^{(t)}}^T G_i^{(t)}]$ is equivalent to minimizing $\textnormal{Tr}( \mathbb{V}\textnormal{ar}_{i \sim p} [G_i^{(t)}])$, the sampling scheme optimizing the convergence speed also minimizes the variance of the stochastic gradient steps performed. The higher the convergence speed, the lower the variance of the stochastic gradient estimate.

The minimization of $\mathbb{E}_{i \sim p} [{G_i^{(t)}}^T G_i^{(t)}]$ is a constrained optimization problem:
\begin{gather*}
    \min_{p} \mathbb{E}_{i \sim p}[{G_i^{(t)}}^T G_i^{(t)}] = \min_{p} \sum_{i=1}^N p_i \|G_i^{(t)}\|_2^2 \quad \textrm{such that } \sum_{i=1}^N p_i = 1 \textrm{ and } \ \forall i \in [1, N], \ p_i > 0 .
\end{gather*}
\begin{flalign*}
&     \textrm{The optimal distribution (solution) is } p^{GN}_i = \|\nabla_\theta \ell(\Psi(\theta,x_i), y_i)\|_2 / \sum_{j=1}^N \|\nabla_\theta \ell(\Psi(\theta,x_j), y_j)\|_2, & &
\end{flalign*}
it is the sampling scheme proportional to the per-sample gradient norms. The proof is in Appendix~\ref{app:init}.

The optimal sampling scheme requires computing $\nabla_\theta \ell(\Psi(\theta,x_i),y_i)$ for all items in the training set, which is too costly to be used in practice. 
Indeed, computing per-sample gradients requires a forward and a backward pass on the whole training data before performing each gradient step.
For this reason, all the works cited in what follows use approximations of the optimal sampling scheme to keep computations tractable.

\subsection{Related work}
\label{subsec:literature}

In this section, we review existing sampling schemes, both in the supervised learning literature and the reinforcement learning one. We start by reviewing the supervised learning literature, where works on importance sampling can be divided into two categories: methods applied to convex problems and methods designed for deep neural networks.

Importance sampling \citep{rubinstein2016simulation} has been a widely studied topic in the context of convex optimization problems in recent years. \citet{bordes2005fast} introduced LASVM, an online algorithm that leverages importance sampling to train kernelized support vector machines. \citet{richtarik2013optimal} subsequently proposed a generalized coordinate descent algorithm that uses importance sampling to optimize the convergence rate of the algorithm. In the context of simple linear classification, the optimal sampling distribution is proportional to the Lipschitz constant of the per-sample loss function \citep{Needell2014SGD, ZhaoZhang2015}. A class of algorithms known as SVRG (Stochastic Variance Reduced Gradient) algorithms \citep{johnson2013accelerating} has been developed to accelerate the convergence of SGD through variance reduction. While these algorithms offer asymptotic improvements, they have been observed to perform worse than SGD with momentum in the multi-modal setting commonly encountered in Deep Learning.

Importance sampling has previously been utilized in the context of deep learning, often in the form of manually tuned sampling schemes. For example, \citet{bengio2009curriculum} manually design a sampling scheme inspired by the way human children learn, while \citet{simo2015discriminative} and \citet{schroff2015facenet} prioritize the sampling of hard examples due to the abundance of easy, non-informative ones. \citet{loshchilov2015online} use the loss to create the sampling distribution, both keeping a history of losses for previously seen samples and sampling proportionally to a loss ranking. It is worth noting the work of \citet{wu2017sampling}, who design a distribution specifically for distance-based losses that maximizes the diversity of losses within a single batch. \citet{fan2017learning} use reinforcement learning to train a neural network that selects samples for another neural network in order to optimize convergence speed. 

\begin{wrapfigure}{r}{0.55\textwidth} 
    \centering
    \includegraphics[width=0.55\textwidth]{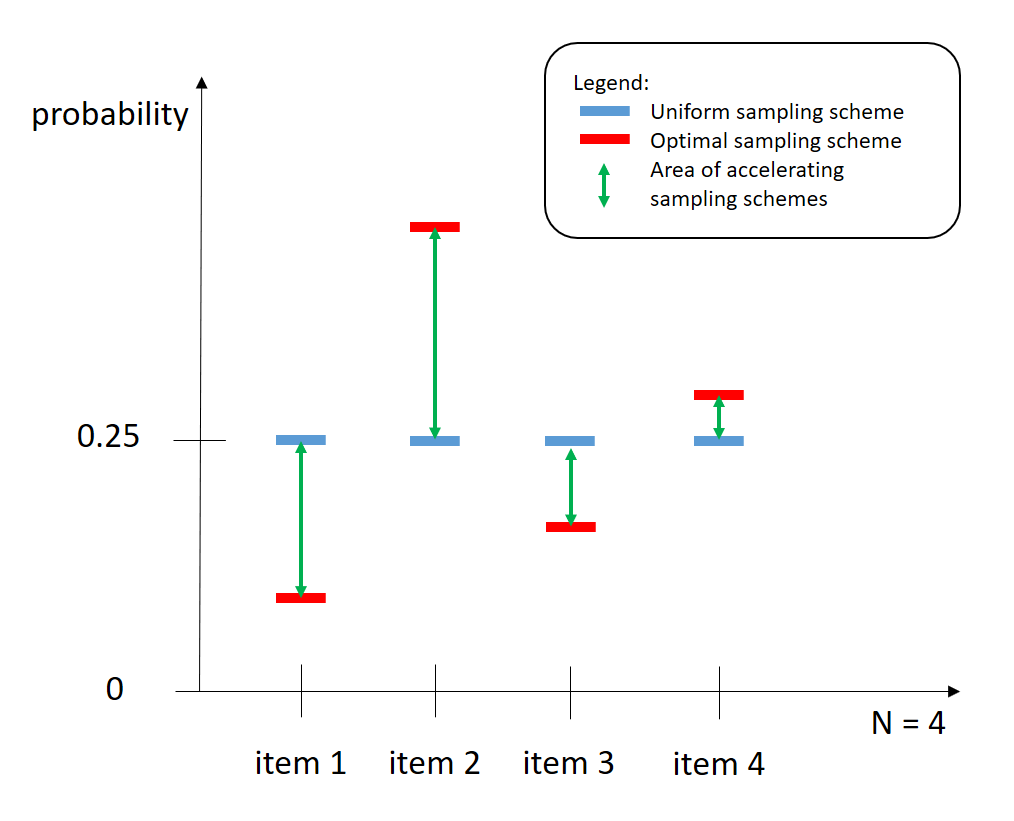}
    \caption{Illustration of the theorem. In this example where $N=4$, the uniform distribution is such that $u_i = 1/4$ and depicted in blue. In red, we draw the ideal sampling scheme. The green arrows show the areas where a sampling scheme is better than the uniform sampling scheme theoretically.}
    \label{fig:sampling}
\end{wrapfigure}

\looseness=-1
The beginning of this section introducing the convergence speed of SGD, and showing that the optimal sampling scheme is proportional to the per-sample gradient norms, is a result of the works done by \citet{Needell2014SGD} and \citet{wang2017accelerating}. All cited works of the previous paragraph uses approximations of the optimal sampling scheme, since it can not be used in practice due to a prohibitive computational cost. Instead of studying approximations, \citet{Alain2016Variance} use clusters of GPU workers to compute the optimal sampling distribution exactly, rendering the computational time acceptable provided high computing resources.

Importance sampling may also be used with deep reinforcement learning algorithms \citep{sutton2018reinforcement}. In particular, off-policy algorithms such as DQN \citep{mnih2015human}, TD3 \citep{fujimoto2018addressing} or SAC \citep{Haarnoja2018SAC} uses a training set (the replay buffer) to train the neural network(s). The first work introducing a non uniform sampling of the replay buffer is Prioritized Experience Replay \citep[PER]{Schaul2015prioritized}. At each iteration, PER samples a mini-batch according to the probability distribution induced by a list of priorities, performs a gradient step and updates the priorities of the selected samples. Hence, PER shares similarities with the work of \citet{loshchilov2015online}, but uses temporal difference errors as priorities instead of per-sample losses. Many improvements have been brought to PER, notably \citep{fujimoto2020equivalence}, \citep{lahire2022large}, \citep{gruslys2018reactor}, and \citep{kumar2020discor}.

\section{Contribution}
\label{sec:contrib}

\subsection{Metric}
\label{subsec:metric}

The first contribution of this work is to provide a metric assessing the quality of a given sampling scheme $p$ compared to the optimal sampling scheme $p^{GN}$ and the uniform sampling scheme $u$ in terms of theoretical convergence speed. Applying directly the optimal sampling scheme is computationally intractable. However, one does not have to apply the best possible sampling scheme to have a theoretical convergence speed better than the convergence speed provided by the uniform sampling scheme. If $p^{GN} \neq u$, there exists a sampling scheme better than $u$ in terms of convergence speed. 

In what follows, we give the necessary conditions for a sampling scheme $p$ to ensure a speed-up compared to the uniform sampling scheme. More mathematically, we give conditions on $p$ so that $S(u) \leq S(p) \leq S(p^{GN})$. 

\textbf{Theorem} For a probability distribution $p$ satisfying $\forall i \in [1, N], \ p_i > 0$, and $\sum_{i=1}^N p_i = 1$, and: 
\begin{equation*}
  \left\{
      \begin{aligned}
      	& p_i \in [u_i; p^{GN}_i] \quad \text{if} \quad u_i \le p^{GN}_i \\
      	& p_i \in [p^{GN}_i ; u_i] \quad \text{if} \quad p^{GN}_i \le u_i, \quad \quad \textnormal{we have $S(u) \leq S(p) \leq S(p^{GN})$.}  \\
      \end{aligned}
    \right.
\end{equation*}

\textbf{Interpretation} Fig. \ref{fig:sampling} illustrates the theorem. A distribution \emph{located between} the ideal sampling $p^{GN}$ and the uniform sampling $u$ yields a theoretical convergence speed higher than the one ensured by the uniform sampling. The closer the sampling scheme $p$ to $p^{GN}$, the higher the convergence speed.

Ideally, when given a new sampling scheme $p$, one has to ensure theoretically that $p$ yields a convergence speed higher than the one obtains with the uniform sampling scheme $u$. However, this theoretical verification might be difficult. If this is impossible, the authors could verify statistically that $p$ is \textit{closer} to $p^{GN}$ than $u$. We propose an algorithm to apply this idea.

Along the optimization process, uniformly sample a set of data (of size $M$) on which you compute the per-sample gradient norms $\| \nabla_\theta \ell(\Psi(\theta, x_j), y_j) \|_2$. Let $\Tilde{p}^{GN}$ be the probability distribution proportional to the per-sample gradient norms on the $M$ selected items: $\Tilde{p}^{GN}_j = \| \nabla_\theta \ell(\Psi(\theta, x_j), y_j) \|_2 / \sum_{k=1}^{M} \| \nabla_\theta \ell(\Psi(\theta, x_k), y_k) \|_2$. On this set of data, compute the given probability distribution $p$. Create the distribution $\Tilde{p}$ over the $M$ items such that $\Tilde{p}_j = p_j / \sum_{k=1}^M p_k$. Let $\Tilde{u}$ be the uniform probability distribution such that $\Tilde{u}_j = 1/M$. Once the three probability distributions $\Tilde{p}^{GN}$, $\Tilde{p}$ and $\Tilde{u}$ have been obtained on the set of size $M$, compute $\mathcal{D}(\Tilde{p}, \Tilde{p}^{GN})$ and $\mathcal{D}(\Tilde{u}, \Tilde{p}^{GN})$, where $\mathcal{D}$ is a metric between two probability distributions, such as the Kullback-Leibler divergence or the Total Variation metric. At each step of the optimization process, $\mathcal{D}(\Tilde{p}, \Tilde{p}^{GN})$ and $\mathcal{D}(\Tilde{u}, \Tilde{p}^{GN})$ can be computed and once the optimization is done, an histogram of the $\mathcal{D}(\Tilde{p}, \Tilde{p}^{GN})$ and $\mathcal{D}(\Tilde{u}, \Tilde{p}^{GN})$ collected can be drawn. If $\mathcal{D}(\Tilde{p}, \Tilde{p}^{GN})$ is, on average, near to zero and far from $\mathcal{D}(\Tilde{u}, \Tilde{p}^{GN})$ which should be larger, then there is evidence in $p$ being close to $p^{GN}$, and $p$ bringing improvement compared to $u$. This algorithm is summarized in Algorithm~\ref{alg:empirical}. Note that, the larger $M$, the better the evaluation is. Indeed, sampling uniformly $M$ items of the training set is a way to capture its diversity without using the whole training set. The larger $M$, the more diverse the selected samples are and the better our method. However, a large $M$ requires many computations: a trade-off given the available computational resources has to be found.

\begin{algorithm}
\caption{Empirical evaluation of a sampling scheme}\label{alg:empirical}
\begin{algorithmic}
\Require $T$ max number of iteration steps, $M$ evaluation set size, $\mathcal{D}$ a metric between probability distributions, $p$ a sampling scheme to assess
\For{$t = 0 ... T$}
\State Select uniformly $M$ samples $(x_j, y_j)$
\State Compute $\| \nabla_\theta \ell(\Psi(\theta, x_j), y_j) \|_2$ for all selected samples
\State Compute $\Tilde{p}^{GN}$ such that $\Tilde{p}^{GN}_j = \| \nabla_\theta \ell(\Psi(\theta, x_j), y_j) \|_2 / \sum_{k=1}^M \| \nabla_\theta \ell(\Psi(\theta, x_k), y_k) \|_2$
\State Compute $p_j$ for all selected samples
\State Compute $\Tilde{p}$ such that $\Tilde{p}_j = p_j / \sum_{k=1}^M p_k$
\State Compute $\Tilde{u}$ such that $\Tilde{u}_j = 1/M$
\State Compute $\mathcal{D}(\Tilde{p}, \Tilde{p}^{GN})$ and $\mathcal{D}(\Tilde{u}, \Tilde{p}^{GN})$ 
\EndFor
\State Draw statistics (e.g. histograms, means...) of the quantities $\mathcal{D}(\Tilde{p}, \Tilde{p}^{GN})$ and $\mathcal{D}(\Tilde{u}, \Tilde{p}^{GN})$  collected along training.
\end{algorithmic}
\end{algorithm}

\subsection{Optimizers and sampling schemes}
\label{subsec:optim}

\looseness=-1
The derivations of \citet{wang2017accelerating} to obtain the optimal sampling scheme in Subsection~\ref{subsec:metric} used the most general form of an SGD step with a mini-batch of one sample, namely $\theta_{t+1} = \theta_t - \eta G_i^{(t)}$, $i$ being the selected index in the training set. 
This yields the optimal sampling scheme being proportional to the per-sample gradient norms.
This subsection takes into account that the standard SGD is no longer used as optimizer. 
Improved versions are now used, such as SGD with momentum, RMSProp or ADAM. Our second contribution is to highlight the differences between these optimizers in terms of analytical expression of the optimal sampling scheme. 
$p^{GN}$ remains the optimal sampling scheme for SGD with momentum, but it is not the case for RMSProp and ADAM.

\textbf{SGD with momentum}

The update equation for SGD with momentum with $B=1$ is $\theta_{t+1} = \theta_t - \eta v_{t} \textrm{ with }  v_{t} = \mu v_{t-1} + G_i^{(t)}$, where $\eta$ is the learning rate and $\mu$ is the momentum coefficient. For initialization, $v_{-1} = 0$ is classically chosen. The derivations for the convergence speed $S$ given the update equations of SGD with momentum yield:
\begin{align*}
S(p) &= - \eta^2 \mu^2 v_{t-1}^T v_{t-1} - 2\eta^2 \mu v_{t-1}^T \mathbb{E}_{i \sim p} \left[ G_i^{(t)} \right] - \eta^2 \mathbb{E}_{i \sim p} \left[ {G_i^{(t)}}^T G_i^{(t)} \right]  \\
& \qquad + 2 \eta \mu (\theta_t - \theta^*)^T v_{t-1} + 2 \eta (\theta_t - \theta^*)^T \mathbb{E}_{i \sim p} \left[ G_i^{(t)} \right].
\end{align*}

Details of these derivations are given in Appendix~\ref{app:SGDM}. Recall that $v_{t-1} = \mu v_{t-2} + G_i^{(t-1)}$ and $G_i^{(t-1)}$ depends on the sampling scheme at time step $t-1$, which is in the past. This means that $v_t$ does not depend on the current sampling scheme $p$ at time step $t$ that has to be optimized. Hence, $v_t$ is a constant with respect to the expectation. Recall also that $\mathbb{E}_{i \sim p}[G_{i}^{(t)}]$ does not depend on $p$. The optimization problem boils down to optimizing $\mathbb{E}_{i \sim p}[{G_{i}^{(t)}}^T G_{i}^{(t)}]$, which yields the same result as for SGD without momentum. The optimal sampling scheme is the sampling scheme proportional to the per-sample gradient norms when the optimizer is SGD with momentum.

\textbf{RMSProp}
\begin{equation*}
    \textrm{The update equation for RMSProp is: } \theta_{t+1} = \theta_t - \frac{\eta}{\epsilon + \sqrt{v_t}}  G_i^{(t)} \textrm{ with } v_t = \alpha v_{t-1} + (1 - \alpha) \| G_i^{(t)}  \|_2^2,
\end{equation*}
where $\epsilon$ is a small positive constant to avoid the division by 0, $\eta$ is the learning rate and $\alpha$ is the moving average parameter, generally set to $0.99$. For initialization, $v_{-1} = 0$ is classically chosen. The derivations for the convergence speed $S$ given the update equations of RMSProp yield:
\begin{align*}
S(p) = -\mathbb{E}_{i \sim p} \left[ \frac{\eta^2 {G_i^{(t)}}^T G_i^{(t)}}{ \left( \epsilon + \sqrt{\alpha v_{t-1} + (1-\alpha) \| G_i^{(t)}  \|_2^2} \right)^2 }  - \frac{2\eta {G_i^{(t)}}^T (\theta_t - \theta^*)}{\epsilon + \sqrt{\alpha v_{t-1} + (1-\alpha) \| G_i^{(t)}  \|_2^2}}  \right].
\end{align*}

\looseness=-1
Details of these derivations are given in Appendix~\ref{app:RMSP}. Contrarily to the derivations done for SGD without momentum, the two terms have to be optimized since they both depend on $p$ (through $G_i^{(t)}$). Given the form of the function to optimize, the solution of the optimization will depend on $\theta^*$. A sampling scheme depending on $\theta^*$ is impractical since we (obviously) do not have access to this quantity. 

The special case where $\alpha = 1$ deserves our attention, since $\alpha$ is in practice often close to $1$. With the initialization $v_{-1} = 0$ and $\alpha = 1$, note that $\forall t, \ v_t = v_{t-1} = 0$. We end up with the equation:
\begin{equation*}
S(p) = -\mathbb{E}_{i \sim p} \left[ \frac{\eta^2}{\epsilon^2} {G_i^{(t)}}^T G_i^{(t)} - \frac{2\eta}{\epsilon} {G_i^{(t)}}^T (\theta_t - \theta^*) \right],
\end{equation*}
which is the same equation than the one obtained with SGD without momentum, the only difference being the learning rate. Once again, the optimization of $\mathbb{E}_{i \sim p} \left[ {G_i^{(t)}}^T G_i^{(t)} \right]$ has to be done. In this special case, the optimal sampling scheme is the one proportional to the per-sample gradient norms. However, note that using RMSProp with $\alpha = 1$ is useless since it stripes RMSProp of what does its specificity, namely the moving average of past gradients.

\textbf{ADAM}

The update equation for ADAM is $\theta_{t+1} = \theta_t - \eta \frac{\hat{m_t}}{\epsilon + \sqrt{\hat{v_t}}}$ with $\hat{m_t} = m_t / (1 - \beta_1^t)$ and $\hat{v_t} = v_t / (1 - \beta_2^t)$ with $m_t = \beta_1 m_{t-1} + (1 - \beta_1) G_i^{(t)}$ and $v_t = \beta_2 v_{t-1} + (1 - \beta_2) \| G_i^{(t)}  \|_2^2$. $\beta_1$ is generally set to $0.9$ and $\beta_2$ to $0.999$. For initialization, $v_{-1} = 0$ and $m_{-1} = 0$. The derivations for the convergence speed $S$ given the update equations of ADAM yield:
\begin{align*}
S(p) &= -\mathbb{E}_{i \sim p} \left[ \frac{\eta^2}{ \left( \epsilon + \sqrt{\frac{\beta_2 v_{t-1} + (1-\beta_2) \| G_i^{(t)}  \|_2^2 }{1 - \beta_2^t}} \right)^2} \frac{1}{(1-\beta_1^t)^2}  \right. \\
& \qquad \quad \quad \times \left( \beta_1 m_{t-1} + (1 - \beta_1) G_i^{(t)}  \right)^T \left( \beta_1 m_{t-1} + (1 - \beta_1) G_i^{(t)} \right) \\
& \qquad \quad \quad \left. - 2\eta \frac{(\theta_t - \theta^*)^T}{\epsilon + \sqrt{\frac{\beta_2 v_{t-1} + (1-\beta_2) \| G_i^{(t)}  \|_2^2 }{1-\beta_2^t}}} \frac{\left( \beta_1 m_{t-1} + (1 - \beta_1) G_i^{(t)} \right)}{(1-\beta_1^t)^2}  \right].
\end{align*}

Details of these derivations are given in Appendix~\ref{app:ADAM}. The derivations lead us to conclusions similar to the ones done for RMSProp. The two terms have to be optimized and the solution of the optimization will depend on $\theta^*$, which makes the optimal sampling scheme impractical. 

The conclusion of this subsection is the following. The sampling scheme proportional to the per-sample gradient norms is not the optimal sampling scheme for RMSProp and ADAM. Applying this sampling scheme with RMSProp and ADAM as optimizer is not theoretically grounded. There is no theoretical evidence that it will bring improvements over the uniform sampling scheme in terms of convergence speed.

\section{Experimental results}
\label{sec:exp}

In this section, we verify experimentally that 1/ a sampling distribution \textit{located between} the uniform and the optimal sampling schemes used with SGD yields a higher convergence speed than the uniform sampling scheme (which illustrates the metric proposed in Subsection~\ref{subsec:metric}), and 2/ using the sampling scheme proportional to the per-sample gradient norms with RMSProp and ADAM does not necessarily yield improvements over the uniform sampling scheme (which illustrates the conclusion of Subsection~\ref{subsec:optim}).

\begin{figure}
\begin{center}
\includegraphics[width=0.99\textwidth]{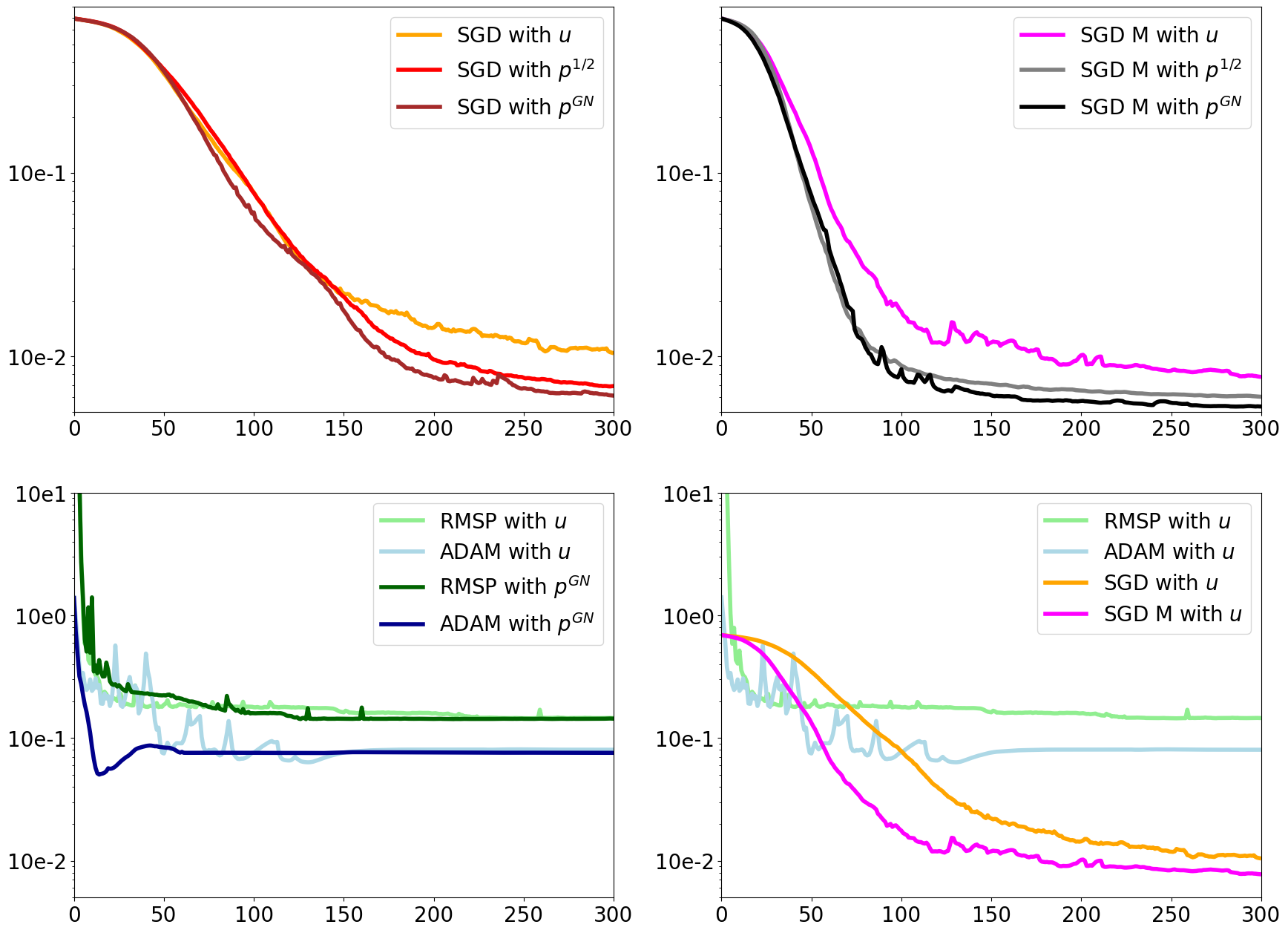}
\caption{Loss decrease for different optimizers used with different sampling schemes. SGD M stands for SGD with momentum, RMSP for RMSProp. The x-axis is the number of gradient steps, the y-axis is the value of the loss. }
\label{fig:results}
\end{center}
\end{figure}

As explained before, the optimal sampling scheme for SGD, proportional to the per-sample gradient norms, cannot be used in practice since it requires computing the individual gradients of all items in the training set. Since it is not possible to evaluate this sampling scheme on well-known data sets that contain too many items, we restrict ourselves to a small part of the well-known data set MNIST \citep{deng2012mnist}, which is under the GNU General Public License v3.0. 

We focus on a binary classification problem, where the goal is to recognize whether the input image is a zero or a one. The training set is equally constituted of images of zero and one, with a total number of items equal to 100. The test set has a size of 2115, and is also equally constituted of images of zeros and ones. We choose a simple convolutional neural network to limit computational burden. Its characteristics are described in Appendix~\ref{app:nn} and the code is available at: \url{https://github.com/thibault-lahire/ImpSampSGD}.  The results were obtained thanks to an Apple Macbook Air 2017: 1,8 GHz Intel Core i5 double.

With a learning rate of $0.01$ and a mini-batch size of $5$, we test the optimizers SGD, RMSProp and ADAM of pytorch \citep{paszke2019pytorch} with default hyper-parameters. For SGD with momentum, we choose a momentum of 0.5. The results reported on Fig.~\ref{fig:results} are an average on 45 runs. The time taken for each run is reported in Table~\ref{table:time}. For a better readability, we only plot the mean results on Fig.~\ref{fig:results}, the error bars around the mean results can be found in Appendix~\ref{app:complete-results}.

Fig.~\ref{fig:results} upper left and right illustrate experimentally that the sampling scheme proportional to the per-sample gradient norms is indeed faster (in terms of iteration) than the uniform sampling scheme for the SGD optimizers (with or without momentum). Fig.~\ref{fig:results} lower left shows it is not the case for RMSProp and ADAM, which corroborates our conclusion of Subsection~\ref{subsec:optim}: the sampling scheme proportional to the per-sample gradient norms is not optimal for RMSProp and ADAM. On the contrary, they do not bring any improvement compared to the uniform sampling scheme.

As an experimental illustration of the metric proposed in Subsection \ref{subsec:metric} for the optimizers SGD and SGD with momentum, we build the sampling scheme $p^{1/2} = (p^{GN}+u)/2$, which is \textit{located between} the uniform sampling schemes $u$ and $p^{GN}$. This probability distribution verifies the hypothesis of our theorem, hence it brings theoretically an improvement on the convergence speed compared to the uniform sampling scheme. The experiments show this improvement is not only theoretical. For SGD and SGD with momentum performed with $p^{1/2}$, the loss decrease is faster than the loss decrease obtained with the uniform sampling, and slower than the one obtained with the optimal sampling scheme (in terms of iterations).

\looseness=-1
Fig.~\ref{fig:results} lower right compares SGD, SGD with momentum, RMSProp and ADAM with the uniform sampling for completeness. Even though this comparison is not the core of our study, the following can be noted. First, RMSProp and ADAM provide a faster loss decrease in the early stages of the optimization process than SGD (with or without momentum). This result is not surprising, as RMSProp and ADAM are known for yielding higher convergence speed than SGD since they use more information: the moving average of gradients helps RMSProp and ADAM driving the optimization towards the optimum \citep{kingma2015ADAM}. However, more surprising is the final result of the optimization: the final loss is lower for SGD than for RMSProp and ADAM. This can be justified by the bias induced by the update equations of RMSProp and ADAM. Whereas SGD minimizes the empirical loss, as proven by \citet{robbins1951stochastic}, it is not the case for RMSProp and ADAM. 

Fig.~\ref{fig:results} cannot be fully appreciated without Table~\ref{table:time}, where the computational time of each sampling scheme used in this work is reported. Whatever the optimizer, the computational time is approximately the same for a given sampling scheme. The computational time for using $p^{GN}$ or $p^{1/2}$ is approximately the same, since $p^{1/2}$ requires $p^{GN}$ to be computed. The more important information of this table is the computational burden of $p^{GN}$ compared to $u$, knowing that the training set is only made of 100 items. Noting that the cost of $p^{GN}$ grows linearly with the size of the training set, this illustrates the fact that $p^{GN}$ is impractical for common deep learning data sets. If the x-axis of Fig.~\ref{fig:results} was the time instead of the number of iterations, the advantage of $p^{GN}$ over $u$ would disappear. Hence, the goal of recent works in importance sampling for deep learning is to find sampling schemes competitive in terms of time which are better than the uniform sampling scheme. 

We now give recommendations for future works dealing with importance sampling and deep learning from a practical point of view. When importance sampling is used, the weighting of samples in the stochastic gradient descent equation must be carefully implemented. Note the crucial importance of the term $1/(Np_i)$ in equation \ref{eq:sgd_non_unif}. If forgotten, the stochastic gradient descent step is performed as if a uniform sampling had been applied (because $1/(Nu_i) = 1$), and this might not lead the optimization process to the minimum. This will lead the optimization process to the minimum of another dataset, which does not exist, where the samples have a distribution depending on $p$. Always from a practical point of view, note also the importance of detaching the weights $1/(Np_i)$ from the computational graph invoked when doing backpropagation. In pytorch \citep{paszke2019pytorch}, for example, the function detach() must be called on the weights $1/(Np_i)$.

\begin{table}
  \caption{Wall-clock time (in seconds) for each optimizer, for one run (300 SGD steps)}
  \label{table:time}
  \centering
  \begin{tabular}{lllll}
    \toprule
    Sampling     & SGD     & SGD M & RMSProp & ADAM \\
    \midrule
    $u$          & 26.5 $\pm$ 1.4   & 24.9 $\pm$ 1.2  & 25.5 $\pm$ 1.6   & 25.7 $\pm$ 1.0   \\
    $p^{GN}$  & 53.0 $\pm$ 1.3   & 56.6 $\pm$ 2.6  & 57.4 $\pm$ 1.6   & 61.4 $\pm$ 4.2 \\
    $p^{1/2}$    & 56.0 $\pm$ 1.5   & 55.3 $\pm$ 1.3  & $\times$   & $\times$ \\
    \bottomrule
  \end{tabular}
\end{table}

\section{Discussion and conclusion}
\label{sec:conclusion}

This work brings two contributions: 1/ a metric to assess the quality of a given sampling scheme and 2/ a study of the interactions between sampling schemes and optimizers. The metric has been verified on a small dataset to keep computations possible. 

The aim of our study was to introduce good practices when dealing with importance sampling for deep learning. Since $p^{GN}$ has a prohibitive cost, approximations of this scheme are proposed and their performance studied in papers dealing with importance sampling for deep learning. These approximations may not be as good as the optimal sampling scheme, but still bring improvements over the uniform sampling scheme. Most of the time, the papers proposing new sampling schemes evaluate them in terms of number of gradient steps performed (x-axis) and loss decrease (y-axis). The good practices our paper introduce are the following. 

First, the comparison with the uniform sampling scheme has also to be done on a computational time basis. The benefits of a given sampling scheme have to be proven in terms of wall-clock time or energy consumption if parallelization is performed. Second, to prevent authors from cherry picking datasets on which the sampling scheme proposed works well, the sampling scheme should perform better than the uniform sampling scheme from a theoretical point of view, verifying the metric proposed in Subsection~\ref{subsec:metric}. Nonetheless, verifying our metric theoretically might be difficult. In this case, the statistical version of our metric proposed in Subsection~\ref{subsec:metric} should at least be verified. 

In this work, we also highlighted in Section~\ref{sec:contrib} that the optimal sampling scheme for RMSProp and ADAM depends on the optimal vector $\theta^*$, which we obviously do not know. As illustrated in the experimental section, using $p^{GN}$ with these optimizers does not yield improvements over the uniform sampling scheme $u$. Nonetheless, this does not mean that $u$ is the optimal sampling scheme for these optimizers, but finding $p$ better than $u$ is even more difficult. Indeed, our metric cannot be applied with these optimizers, even in its statistical version.

\looseness=-1
Overall, given the speed up that can be obtained with optimizers such as RMSProp or ADAM without importance sampling compared to standard SGD, and the fact that importance sampling must be carefully used with these optimizers, developing new sampling schemes filling all conditions (efficiency in wall-clock time, and theoretical guarantee of doing better than uniform sampling) with RMSProp or ADAM appears to us a very difficult but challenging task.

\newpage
\bibliographystyle{apalike}
\bibliography{biblio}


\newpage

\appendix

\newpage
\section{Derivation of the optimal sampling scheme for SGD}
\label{app:init}

In this appendix, we proove that the sampling scheme proportional to the per-sample gradient norms is the optimal sampling scheme for SGD in its most standard form. Following the notations of \citet{wang2017accelerating}, let us recall the convergence speed $S$ of SGD under a sampling scheme $p$ as
\begin{equation*}
    S(p) = -\mathbb{E}_{i \sim p} \left[ \|\theta_{t+1} - \theta^*\|_2^2 - \|\theta_t - \theta^*\|_2^2 \right].
\end{equation*}

We recall that a stochastic gradient descent update has the form $\theta_{t+1} = \theta_t - \eta G_i^{(t)}$, where $G_i^{(t)}$ is the gradient estimate built from sampling element $i$ with probability $p$.
The following derivations from \citep{wang2017accelerating} yield:
\begin{align*}
S(p) &= -\mathbb{E}_{i \sim p} \left[ \|\theta_{t+1} - \theta^*\|_2^2 - \|\theta_t - \theta^*\|_2^2 \right] \\
&= -\mathbb{E}_{i \sim p} \left[ \theta_{t+1}^T \theta_{t+1} - 2 \theta_{t+1}^T \theta^* -  \theta_{t}^T \theta_{t} +2 \theta_{t}^T \theta^* \right] \\
&= -\mathbb{E}_{i \sim p} \left[ (\theta_t - \eta G_i^{(t)})^T (\theta_t - \eta G_i^{(t)}) + 2\eta {G_i^{(t)}}^T \theta^* - \theta_t^T \theta_t \right] \\
&= -\mathbb{E}_{i \sim p} \left[ -2 \eta (\theta_t - \theta^*)^T G_i^{(t)} + \eta^2 {G_i^{(t)}}^T G_i^{(t)} \right] \\
&= 2 \eta (\theta_t - \theta^*)^T \mathbb{E}_{i \sim p}[G_i^{(t)}] - \eta^2 \mathbb{E}_{i \sim p} [{G_i^{(t)}}^T G_i^{(t)}].
\end{align*}
Recall also from Eq.~\ref{eq:const} that $\mathbb{E}_{i \sim p}[G_{i}^{(t)}]$ is a constant with respect to $p$. Hence, it is possible to gain a speed-up by sampling from the distribution that minimizes $\mathbb{E}_{i \sim p} [{G_i^{(t)}}^T G_i^{(t)}]$. 

The minimization of $\mathbb{E}_{i \sim p} [{G_i^{(t)}}^T G_i^{(t)}]$ is a constrained optimization problem:
\begin{gather*}
    \min_{p} \mathbb{E}_{i \sim p}[{G_i^{(t)}}^T G_i^{(t)}] = \min_{p} \sum_{i=1}^N p_i \|G_i^{(t)}\|_2^2 \quad \textrm{such that } \sum_{i=1}^N p_i = 1 \textrm{ and } \ \forall i \in [1, N],  p_i > 0 .
\end{gather*}
Recall that $G_i^{(t)} = w_i \nabla_{\theta} \ell(\Psi(\theta_t, x_i), y_i)$ and $w_i = 1/(Np_i)$. Let $g_i = \| \nabla_\theta \ell (\Psi(\theta_t, x_i), y_i)\|_2$. The problem boils down to:
\begin{gather*}
     \min_p \frac{1}{N^2} \sum_{i=1}^N \frac{1}{p_i} g_i^2, \quad
     \textrm{such that } \sum_{i=1}^N p_i = 1 \textrm{ and } \forall i \in [1, N], p_i > 0.
\end{gather*}
\begin{flalign*}
&     \textrm{The optimal distribution is } p^{GN}_i = \|\nabla_\theta \ell(\Psi(\theta,x_i), y_i)\|_2 / \sum_{j=1}^N \|\nabla_\theta \ell(\Psi(\theta,x_j), y_j)\|_2, & &
\end{flalign*}
it is the sampling scheme proportional to the per-sample gradient norms.

\begin{proof}
We note $\mu \in \mathbb{R}$ the Lagrange multiplier associated to the equality constraint, $\nu \in \mathbb{R}_+^N$ the Lagrange multipliers associated to the inequality constraints. Hence:
\begin{equation*}
    \textnormal{Lag}(p, \mu, \nu) = \sum_{i=1}^N \frac{1}{p_i} g_i^2 + \mu \left(\sum_{i=1}^N p_i-1 \right) -\sum_{i=1}^N \nu_i p_i
\end{equation*}
Setting the derivatives of the Lagrangian with respect to the primal variables yields: 
\begin{equation*}
    \forall \ i \in [ 1, N ], \ -\frac{g_i^2}{p_i^2} + \mu - \nu_i = 0
\end{equation*}
Multiplying the above equation by $p_i$ and using $\forall \ i, \ p_i\nu_i = 0$ (complementary slackness), we have: $p_i = g_i / \sqrt{\mu}$, which yields the result.
\end{proof}

\newpage
\section{Proof of the theorem associated to the metric}
\label{app:prooftheorem}

One contribution of this work is to provide a metric assessing the quality of a given sampling scheme $p$ compared to the optimal sampling scheme $p^{GN}$ and the uniform sampling scheme $u$. Applying directly the optimal sampling scheme is computationally intractable. However, one does not have to apply the best possible sampling scheme to have a theoretical convergence speed higher than the convergence speed provided by the uniform sampling scheme. If $p^{GN} \neq u$, there exists a sampling scheme better than $u$ in terms of convergence speed. 

Let's recall that 
\begin{align*}
    S(p) &= 2 \eta (\theta_t - \theta^*)^T \mathbb{E}_{i \sim p} \left[ G_i^{(t)} \right] - \eta^2 \mathbb{E}_{i \sim p} \left[ {G_i^{(t)}}^T G_i^{(t)} \right] \\
    &= 2 \eta (\theta_t - \theta^*)^T \frac{1}{N} \sum_{i=1}^N \nabla_\theta \ell(\Psi(\theta_t, x_i), y_i) - \eta^2 \sum_{i=1}^N \frac{1}{p_i} \nabla_\theta \ell(\Psi(\theta_t, x_i), y_i)^T \nabla_\theta \ell(\Psi(\theta_t, x_i), y_i) \\
    &= \textrm{constant} - \frac{\eta^2}{N^2} \sum_{i=1}^N \frac{1}{p_i} \| \nabla_\theta \ell(\Psi(\theta_t, x_i), y_i) \|_2^2.
\end{align*}
Let's introduce $H(p) = \sum_{i=1}^N \frac{1}{p_i} \| \nabla_\theta \ell(\Psi(\theta_t, x_i), y_i) \|_2^2$.

In what follows, we give the necessary conditions for a sampling scheme $p$ to ensure a speed-up compared to the uniform sampling scheme. More mathematically, we give conditions on $p$ so that $S(u) \leq S(p) \leq S(p^{GN})$, or $H(p^{GN}) \leq H(p) \leq H(u)$. 

\textbf{Theorem} For a discrete probability distribution $p$ satisfying $\sum_{i=1}^N p_i = 1$, and, $\forall i \in [1, N], \ p_i > 0$, and : 
\begin{equation*}
  \left\{
      \begin{aligned}
      	p_i \in [u_i; p^{GN}_i] \quad \text{if} \quad u_i \le p^{GN}_i \\
      	p_i \in [p^{GN}_i ; u_i] \quad \text{if} \quad p^{GN}_i \le u_i, \\
      \end{aligned}
    \right.
\end{equation*}
$\quad $ we have $H(p^{GN}) \le H(p) \le H(u)$, which is equivalent to $S(p^{GN}) \ge S(p) \ge S(u)$.  \\

\textbf{Proof} Since $p^{GN}$ is the optimal sampling scheme, yielding the highest convergence speed by definition, $S(p^{GN}) \ge S(p)$ and $H(p^{GN}) \le H(p)$. Let's introduce $\forall i \in [1, N], \ t_i \in [0;1]$ such that $p_i = t_i u_i + (1-t_i)p_i^{GN}$. It yields $u_i - p_i = (1-t_i)(u_i-p_i^{GN})$. Hence:

\begin{align*}
H(p) - H(u) &= \sum_{i=1}^N \left(\frac{1}{p_i} - \frac{1}{u_i}\right) ||\nabla_\theta \ell(\Psi(\theta,x_i), y_i)||_2^2 \\
&= \sum_{i=1}^N \left(\frac{u_i - p_i}{p_i u_i}\right)||\nabla_\theta \ell(\Psi(\theta,x_i), y_i)||_2^2 \\
&= \sum_{i=1}^N (1-t_i) \frac{u_i - p_i^{GN}}{u_i p_i^{GN}} \frac{u_i p_i^{GN}}{u_i p_i} ||\nabla_\theta \ell(\Psi(\theta,x_i), y_i)||_2^2 \\
&= \sum_{i=1}^N \left( \frac{1}{p_i^{GN}} - \frac{1}{u_i} \right) ||\nabla_\theta \ell(\Psi(\theta,x_i), y_i)||_2^2 \times (1-t_i)\frac{p_i^{GN}}{p_i} \\
&\le \sum_{i=1}^N \left( \frac{1}{p_i^{GN}} - \frac{1}{u_i} \right) ||\nabla_\theta \ell(\Psi(\theta,x_i), y_i)||_2^2 \times \max_{1 \le j \le N} \left( (1-t_j) \frac{p_j^{GN}}{p_j} \right) \\
&\le (H(p^{GN}) - H(u)) M, \quad \text{with} \quad M = \max_{1 \le j \le N} \left( (1-t_j) \frac{p_j^{GN}}{p_j} \right) \\
&\le 0
\end{align*}

where the last inequality is justified by $M \ge 0$ and $H(p^{GN}) - H(u) \le 0$. This proves that $H(p) \le H(u)$ and $S(p) \ge S(u)$. 
$\hfill \blacksquare$

\section{Convergence speed for SGD with momentum}
\label{app:SGDM}

The update equation for SGD with momentum with $B=1$ is $\theta_{t+1} = \theta_t - \eta v_{t} \textrm{ with }  v_{t} = \mu v_{t-1} + G_i^{(t)}$, where $\eta$ is the learning rate and $\mu$ is the momentum coefficient. For initialization, $v_{-1} = 0$ is classically chosen.

We write the derivations for the convergence speed $S$ given the update equations of SGD with momentum. It yields:

\begin{align*}
S(p) &= -\mathbb{E}_{i \sim p} \left[ \|\theta_{t+1} - \theta^*\|_2^2 - \|\theta_t - \theta^*\|_2^2 \right] \\
&= -\mathbb{E}_{i \sim p} \left[ \theta_{t+1}^T \theta_{t+1} - 2 \theta_{t+1}^T \theta^* -  \theta_{t}^T \theta_{t} +2 \theta_{t}^T \theta^* \right] \\
&= -\mathbb{E}_{i \sim p} \left[ (\theta_t - \eta v_t)^T (\theta_t - \eta v_t) - 2 (\theta_t - \eta v_t)^T \theta^* - \theta_t^T \theta_t + 2 \theta_t^T \theta^* \right] \\
&= -\mathbb{E}_{i \sim p} \left[ \eta^2 v_t^T v_t - 2 \eta (\theta_t - \theta^*)^T v_t\right] \\
&= -\mathbb{E}_{i \sim p} \left[ \eta^2 \left( \mu v_{t-1} + G_i^{(t)} \right)^T \left( \mu v_{t-1} + G_i^{(t)} \right) - 2 \eta (\theta_t - \theta^*)^T \left( \mu v_{t-1} + G_i^{(t)} \right) \right] \\
&= -\mathbb{E}_{i \sim p} \left[ \eta^2 \mu^2 v_{t-1}^T v_{t-1} + 2\eta^2 \mu v_{t-1}^T G_i^{(t)} + \eta^2  {G_i^{(t)}}^T G_i^{(t)} - 2 \eta(\theta_t - \theta^*)^T \left( \mu v_{t-1} + G_i^{(t)} \right)   \right] \\
&= - \eta^2 \mu^2 v_{t-1}^T v_{t-1} - 2\eta^2 \mu v_{t-1}^T \mathbb{E}_{i \sim p} \left[ G_i^{(t)} \right] - \eta^2 \mathbb{E}_{i \sim p} \left[ {G_i^{(t)}}^T G_i^{(t)} \right]  \\
& \qquad + 2 \eta \mu (\theta_t - \theta^*)^T v_{t-1} + 2 \eta (\theta_t - \theta^*)^T \mathbb{E}_{i \sim p} \left[ G_i^{(t)} \right].
\end{align*}

\newpage
\section{Convergence speed for RMSProp}
\label{app:RMSP}

\begin{equation*}
    \textrm{The update equation for RMSProp is: } \theta_{t+1} = \theta_t - \frac{\eta}{\epsilon + \sqrt{v_t}}  G_i^{(t)} \textrm{ with } v_t = \alpha v_{t-1} + (1 - \alpha) \| G_i^{(t)}  \|_2^2
\end{equation*}
where $\epsilon$ is a small positive constant to avoid the division by 0, $\eta$ is the learning rate and $\alpha$ is the moving average parameter, generally set to $0.99$. For initialization, $v_{-1} = 0$ is classically chosen.

We write the derivations for the convergence speed $S$ given the update equations of RMSProp. It yields:

\begin{align*}
S(p) &= -\mathbb{E}_{i \sim p} \left[ \|\theta_{t+1} - \theta^*\|_2^2 - \|\theta_t - \theta^*\|_2^2 \right] \\
&= -\mathbb{E}_{i \sim p} \left[ \theta_{t+1}^T \theta_{t+1} - 2 \theta_{t+1}^T \theta^* -  \theta_{t}^T \theta_{t} +2 \theta_{t}^T \theta^* \right] \\
&= -\mathbb{E}_{i \sim p} \left[ \left( \theta_t - \frac{\eta}{\epsilon + \sqrt{\alpha v_{t-1} + (1-\alpha) \| G_i^{(t)}  \|_2^2}} {G_{i}^{(t)}} \right)^T \right. \\
& \left. \qquad \qquad \times \left( \theta_t - \frac{\eta}{\epsilon + \sqrt{\alpha v_{t-1} + (1-\alpha) \| G_i^{(t)}  \|_2^2}} {G_{i}^{(t)}} \right) \right. \\
& \left. \qquad \qquad  - 2 \left( \theta_t - \frac{\eta}{\epsilon + \sqrt{\alpha v_{t-1} + (1-\alpha) \| G_i^{(t)}  \|_2^2}} {G_{i}^{(t)}} \right)^T \theta^* -\theta_t^T \theta_t + 2 \theta_t^T \theta^* \right] \\
&= -\mathbb{E}_{i \sim p} \left[ \frac{\eta^2}{ \left( \epsilon + \sqrt{\alpha v_{t-1} + (1-\alpha) \| G_i^{(t)}  \|_2^2 } \right)^2} {G_i^{(t)}}^T G_i^{(t)} \right. \\
& \left. \qquad - \frac{2\eta}{\epsilon + \sqrt{\alpha v_{t-1} + (1-\alpha) \| G_i^{(t)}  \|_2^2}} {G_{i}^{(t)}}^T \theta_t + \frac{2\eta}{\epsilon + \sqrt{\alpha v_{t-1} + (1-\alpha) \| G_i^{(t)}  \|_2^2}} {G_{i}^{(t)}}^T \theta^* \right] \\
&= -\mathbb{E}_{i \sim p} \left[ \frac{\eta^2 {G_i^{(t)}}^T G_i^{(t)}}{ \left( \epsilon + \sqrt{\alpha v_{t-1} + (1-\alpha) \| G_i^{(t)}  \|_2^2} \right)^2 }  - \frac{2\eta {G_i^{(t)}}^T (\theta_t - \theta^*)}{\epsilon + \sqrt{\alpha v_{t-1} + (1-\alpha) \| G_i^{(t)}  \|_2^2}}  \right].
\end{align*}

\newpage
\section{Convergence speed for ADAM}
\label{app:ADAM}

The update equation for ADAM is $\theta_{t+1} = \theta_t - \eta \frac{\hat{m_t}}{\epsilon + \sqrt{\hat{v_t}}}$ with $\hat{m_t} = m_t / (1 - \beta_1^t)$ and $\hat{v_t} = v_t / (1 - \beta_2^t)$ with $m_t = \beta_1 m_{t-1} + (1 - \beta_1) G_i^{(t)}$ and $v_t = \beta_2 v_{t-1} + (1 - \beta_2) \| G_i^{(t)}  \|_2^2$. $\beta_1$ is generally set to $0.9$ and $\beta_2$ to $0.999$. For initialization, $v_{-1} = 0$ and $m_{-1} = 0$ are classically chosen.

We write the derivations for the convergence speed $S$ given the update equations of ADAM. It yields:

\begin{align*}
S(p) &= -\mathbb{E}_{i \sim p} \left[ \|\theta_{t+1} - \theta^*\|_2^2 - \|\theta_t - \theta^*\|_2^2 \right] \\
&= -\mathbb{E}_{i \sim p} \left[ \theta_{t+1}^T \theta_{t+1} - 2 \theta_{t+1}^T \theta^* -  \theta_{t}^T \theta_{t} +2 \theta_{t}^T \theta^* \right] \\
&= -\mathbb{E}_{i \sim p} \left[ \left( \theta_t - \eta \frac{\hat{m_t}}{\epsilon + \sqrt{\hat{v_t}}} \right)^T \left( \theta_t - \eta \frac{\hat{m_t}}{\epsilon + \sqrt{\hat{v_t}}} \right) - 2 \left( \theta_t - \eta \frac{\hat{m_t}}{\epsilon + \sqrt{\hat{v_t}}} \right)^T  \theta^* \right. \\
& \left. \qquad \qquad \quad - \theta_t^T \theta_t + 2 \theta_t^T \theta^* \right] \\
&= -\mathbb{E}_{i \sim p} \left[ \eta^2 \frac{\hat{m_t}^T \hat{m_t}}{(\epsilon + \sqrt{\hat{v_t}})^2} - 2\eta \frac{\hat{m_t}^T \theta_t}{\epsilon + \sqrt{v_t}} + 2\eta \frac{\hat{m_t}^T \theta^*}{\epsilon + \sqrt{v_t}} \right] \\
&= -\mathbb{E}_{i \sim p} \left[ \eta^2 \frac{\hat{m_t}^T \hat{m_t}}{(\epsilon + \sqrt{\hat{v_t}})^2} - 2\eta (\theta_t - \theta^*)^T \frac{\hat{m_t}}{\epsilon + \sqrt{v_t}} \right] \\
&= -\mathbb{E}_{i \sim p} \left[ \frac{\eta^2}{ \left( \epsilon + \sqrt{\frac{v_t}{1 - \beta_2^t}} \right)^2} \frac{m_t^T m_t}{ \left( 1-\beta_1^t \right)^2} - 2\eta \frac{(\theta_t - \theta^*)^T}{\epsilon + \sqrt{\frac{v_t}{1-\beta_2^t}}} \frac{m_t}{1 - \beta_1^t} \right] \\
&= -\mathbb{E}_{i \sim p} \left[ \frac{\eta^2}{ \left( \epsilon + \sqrt{\frac{\beta_2 v_{t-1} + (1-\beta_2) \| G_i^{(t)}  \|_2^2 }{1 - \beta_2^t}} \right)^2} \frac{1}{(1-\beta_1^t)^2}  \right. \\
& \qquad \qquad \quad \times \left( \beta_1 m_{t-1} + (1 - \beta_1) G_i^{(t)}  \right)^T \left( \beta_1 m_{t-1} + (1 - \beta_1) G_i^{(t)} \right) \\
& \qquad \qquad \quad \left. - 2\eta \frac{(\theta_t - \theta^*)^T}{\epsilon + \sqrt{\frac{\beta_2 v_{t-1} + (1-\beta_2) \| G_i^{(t)}  \|_2^2 }{1-\beta_2^t}}} \frac{\left( \beta_1 m_{t-1} + (1 - \beta_1) G_i^{(t)} \right)}{(1-\beta_1^t)^2}  \right]. 
\end{align*}

\section{Neural network architecture}
\label{app:nn}

Our task is a binary classification on the MNIST dataset, and this dataset does not require complex architectures to obtain an acceptable classifier. For this reason, and to limit the computational cost, we choose a simple convolutional neural network. The first layer is a 2D convolutional layer with an input channel size of 1, output channel size of 5, a kernel size of 5 and a stride of 1. The second layer is a 2D convolutional layer with an input channel size of 5, output channel size of 10, a kernel size of 5 and a stride of 1. The third layer is a fully connected layer taking a vector of size 4000 as input and returning a vector of size 100. The last layer is a fully connected layer taking a vector of size 100 as input and returning a vector of size 2, since our problem is a binary classification. The activation function at the output of each layer, except the last one, is the ReLU (Rectified Linear Unit). The neural network returns the log softmax of the output of the last layer. 

\section{Additional experimental results}
\label{app:complete-results}

\begin{figure}[H]
\begin{center}
\includegraphics[width=0.99\textwidth]{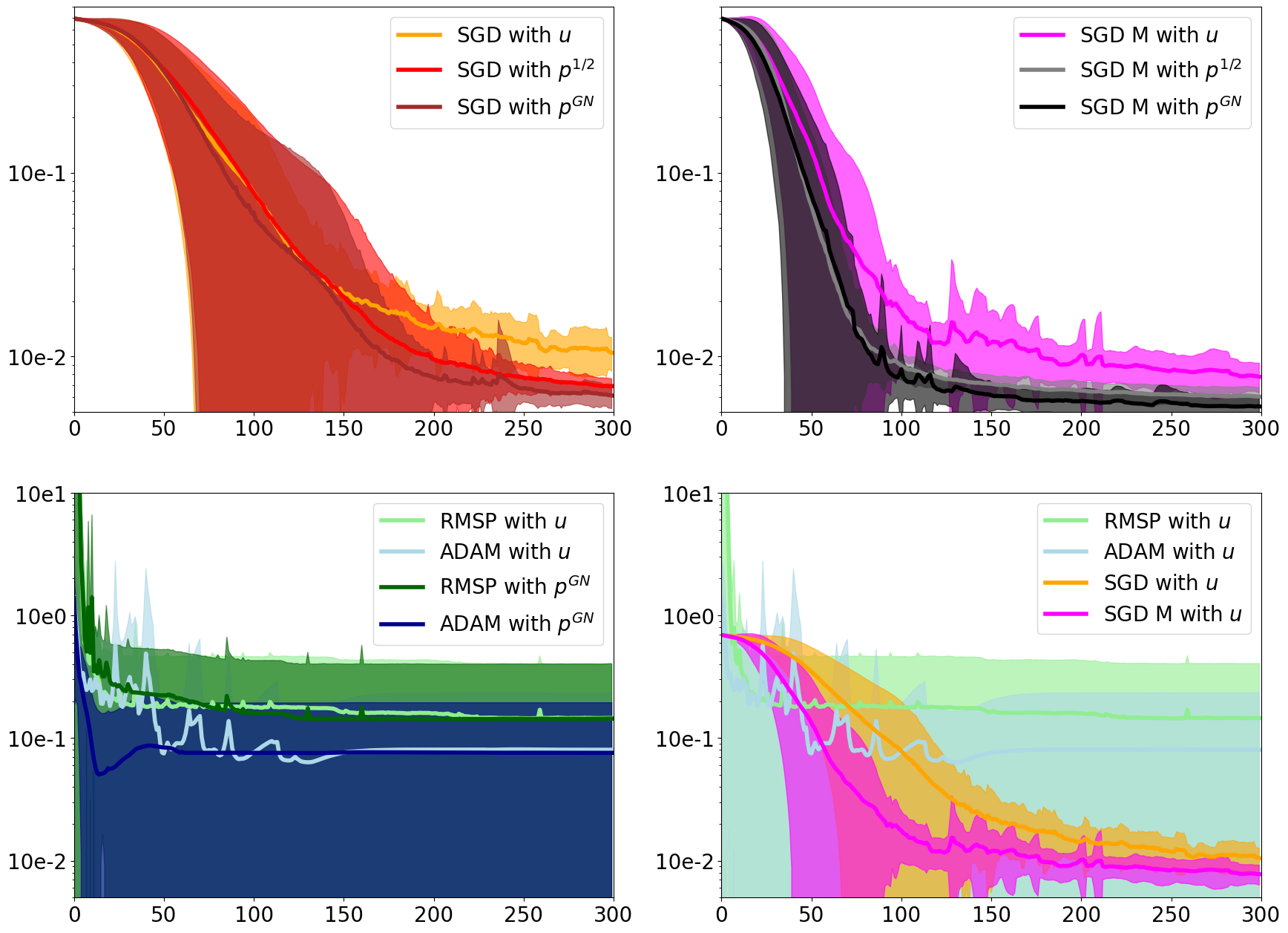}
\caption{Loss decrease for different optimizers used with different sampling schemes. SGD M stands for SGD with momentum, RMSP for RMSProp. The x-axis is the number of gradient steps, the y-axis is the value of the loss. }
\label{fig:results_complete}
\end{center}
\end{figure}

In this appendix can be found the experimental results with error bars (Fig.~\ref{fig:results_complete}). Let $\mu$ be the mean result and $\sigma$ the standard deviation. The colorized area around $\mu$ is $[\mu-\sigma; \mu+\sigma]$.

\end{document}